\documentclass[letterpaper, 10 pt, conference]{ieeeconf}  

\IEEEoverridecommandlockouts                              

\usepackage{algorithm}
\usepackage[noend]{algpseudocode}
\usepackage{amssymb}
\usepackage{mathtools}

\newtheorem{theorem}{Theorem}

\usepackage{booktabs}
\newcommand{\para}[1]{\smallskip \noindent \textit{#1}}

\input

\newcommand{\allTk}[0]{Z}
\newcommand{\tk}[1]{z_{#1}}
\newcommand{\allP}[0]{P}
\newcommand{\p}[1]{p_{#1}}
\newcommand{\inTk}[0]{\textsc{inTk}}
\newcommand{\numIn}[0]{\textsc{numIn}}
\newcommand{\outTk}[0]{\textsc{outTk}}
\newcommand{\numOut}[0]{\textsc{numOut}}
\newcommand{\outP}[0]{\textsc{outP}}

\newcommand{\allM}[0]{M}
\newcommand{\m}[1]{m_{#1}}
\newcommand{\mProc}[0]{\mathcal{P}}
\newcommand{\runtime}[0]{\textsc{runtime}}
\newcommand{\bInM}[0]{\boldsymbol{\mathcal{I}}}
\newcommand{\bIn}[0]{\mathcal{I}}
\newcommand{\bOutM}[0]{\boldsymbol{\mathcal{O}}}
\newcommand{\bOut}[0]{\mathcal{O}}
\newcommand{\srcM}[0]{M_{src}}
\newcommand{\sinkM}[0]{M_{sink}}
\newcommand{\inputOn}[0]{\textsc{inputOn}}
\newcommand{\outputOn}[0]{\textsc{outputOn}}

\newcommand{\allCl}[0]{C}
\newcommand{\cl}[1]{c_{#1}}

\newcommand{\layoutG}[0]{G_L}
\newcommand{\layoutE}[0]{E_L}
\newcommand{\inputCl}[0]{\textsc{inputCell}}

\newcommand{\ag}[1]{a_{#1}}
\newcommand{\allAg}[0]{A}
\newcommand{\agClM}[0]{\boldsymbol{\agCl{}}}
\newcommand{\agCl}[0]{\pi}
\newcommand{\agTkM}[0]{\boldsymbol{\agTk{}}}
\newcommand{\agTk}[0]{\sigma}
\newcommand{\nullTk}[0]{\tk{0}}

\newcommand{\road}[1]{R_{#1}}
\newcommand{\rLen}[0]{\textsc{len}}
\newcommand{\allRoads}[0]{\mathcal{R}}
\newcommand{\junction}[1]{J_{#1}}
\newcommand{\allJunctions}[0]{\mathcal{J}}
\newcommand{\entry}[0]{\textsc{entry}}
\newcommand{\exit}[0]{\textsc{exit}}

\newcommand{\assignM}[0]{\boldsymbol{\assign{}}}
\newcommand{\assign}[0]{\alpha}
\newcommand{\rateM}[0]{\boldsymbol{\rate{}}}
\newcommand{\rate}[0]{\lambda}
\newcommand{\tpGen}[0]{\mathcal{G}}
\newcommand{\sfState}[0]{\phi}

\newcommand{\throughput}[0]{\theta}

\newcommand*\circled[1]{\raisebox{.5pt}{\textcircled{\raisebox{-.9pt} {#1}}}}

\newcommand{\TSMILP}[0]{TS-MILP}

\newcommand{\inBM}[0]{\boldsymbol{\inB{}}}
\newcommand{\inB}[0]{inB}
\newcommand{\outBM}[0]{\boldsymbol{\outB{}}}
\newcommand{\outB}[0]{outB}
\newcommand{\pkM}[0]{\boldsymbol{\pk{}}}
\newcommand{\pk}[0]{pk}
\newcommand{\dpM}[0]{\boldsymbol{\dpF{}}}
\newcommand{\dpF}[0]{dp}
\newcommand{\T}[0]{T}
\newcommand{\numEpochs}[0]{N_e}
\newcommand{\epochLen}[0]{T_e}
\newcommand{\totalInB}[0]{totInB}
\newcommand{\totalOutB}[0]{totOutB}

\newcommand{\tsPlanner}[0]{\textsc{tsPlanner}}
\newcommand{\inst}[0]{inst}
\newcommand{\timer}[0]{timer}
\newcommand{\sol}[0]{sol}
\newcommand{\planForNumEpochs}[0]{\textsc{planForNumEpochs}}
\newcommand{\maxFailureNum}[0]{\gamma}
\newcommand{\incrEpochLen}[0]{\delta}
\newcommand{\runsSinceBestSol}[0]{runsSinceBestSol}
\newcommand{\solveTSILP}[0]{\textsc{runMILP}}

\newcommand{\TSACES}[0]{\textsc{TS-ACES}}
\newcommand{\TSTGEN}[0]{\mathcal{G}_{TS}}
\newcommand{\atM}[0]{\boldsymbol{at}}
\newcommand{\at}[0]{at}
\newcommand{\canChgTk}[0]{canChgTk}
\newcommand{\arrivalT}[0]{enT}
\newcommand{\genOccupancyVec}[0]{\textsc{genOccupancyVec}}
\newcommand{\adjustStateForNewEpoch}[0]{\textsc{adjustStateForNewEpoch}}
\newcommand{\moveAgentOnJunction}[0]{\textsc{moveAgentOnJunction}}
\newcommand{\moveAgentsOnRoad}[0]{\textsc{moveAgentsOnRoad}}

\newcommand{\agentAt}[0]{at}
\newcommand{\cell}[0]{\textsc{cell}}
\newcommand{\nullVal}[0]{\textsc{null}}
\newcommand{\rndEle}[0]{\textsc{rndEle}}
\newcommand{\reversePath}[0]{\textsc{reversePath}}

\newcommand{\head}[0]{\textsc{head}}
\newcommand{\tail}[0]{\textsc{tail}}
\newcommand{\depositToken}[0]{\textsc{depositToken}}
\newcommand{\pickupToken}[0]{\textsc{pickupToken}}
\newcommand{\keepToken}[0]{\textsc{keepToken}}
\newcommand{\genPositionVec}[0]{\textsc{genPositionVec}}
\newcommand{\isSourceM}[0]{\textsc{isSourceM}}

\newcommand{\initializeSF}[0]{\textsc{initializeSF}}

\newcommand{\singleAgentPlan}[0]{\zeta}

\title{\LARGE \bf
An Anytime, Scalable and Complete Algorithm for Embedding a Manufacturing Procedure in a Smart Factory
}

\author{Christopher Leet$^{1}$, Aidan Sciortino$^{2}$, and Sven Koenig$^{3}$
\thanks{$^{1}$University of Southern California
        {\tt\small cjleet@usc.edu}}%
\thanks{$^{2}$University of Rochester
        {\tt\small asciorti@u.rochester.edu}}%
\thanks{$^{3}$University of Southern California
        {\tt\small skoenig@usc.edu}}%
}

\begin{document}

\maketitle
\thispagestyle{empty}
\pagestyle{empty}

\begin{abstract}
Modern automated factories increasingly run manufacturing procedures using a matrix of programmable machines, such as 3D printers, interconnected by a programmable transport system, such as a fleet of tabletop robots. To embed a manufacturing procedure into a smart factory, an operator must: (a) assign each of its processes to a machine and (b) specify how agents should transport parts between machines. The problem of embedding a manufacturing process into a smart factory is termed the Smart Factory Embedding (SFE) problem. State-of-the-art  SFE solvers can only scale to factories containing a couple dozen machines. Modern smart factories, however, may contain hundreds of machines. We fill this hole by introducing the first highly scalable solution to the SFE, \TSACES{}, the Traffic System based Anytime Cyclic Embedding Solver. We show that \TSACES{} is complete and can scale to SFE instances based on real industrial scenarios with more than a hundred machines.
\end{abstract}

\begin{figure*}[t]
    \centering\includegraphics[width=\linewidth, trim={0 0 0 0},clip]{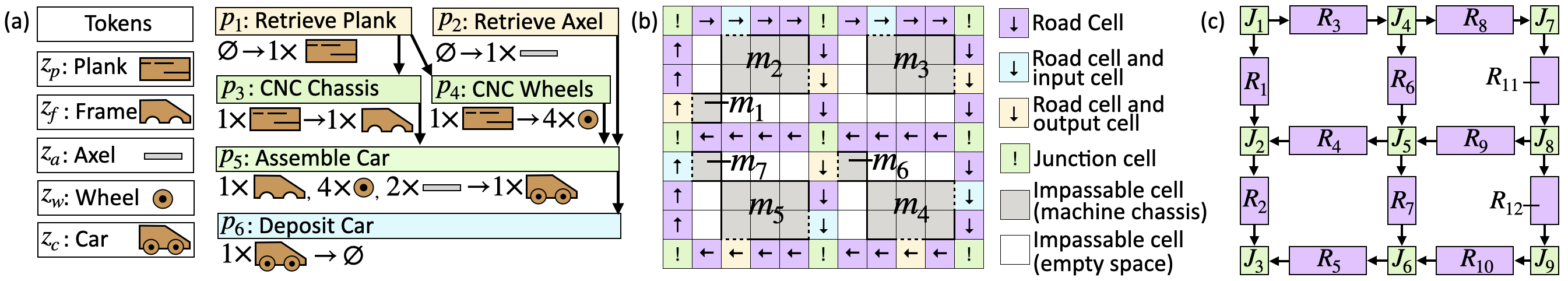}

    \vspace{-3mm}
    \caption{(a) An example manufacturing procedure. Source and sink process are yellow and blue. (b) An example smart factory and (c) its traffic system.}
    \vspace{-5mm}
    \label{fig:example_problem}
\end{figure*}

\section{INTRODUCTION}

Flexible manufacturing is a key objective of the modern manufacturing industry~\cite{enabling_flexible_manufacturing}. A smart factory is flexible if it can be easily reconfigured to produce different products. Flexible manufacturing systems can reduce the cost of producing new products, lower the time required to fulfill orders, and allow products to be customized.  To perform flexible manufacturing, a smart factory needs two components:


\para{Flexible Machines.} Flexible machines are general purpose machines such as CNC machines which can be used to perform a range of manufacturing processes. Flexible machines can be easily reprogrammed with a new process when a smart factory's manufacturing procedure changes. 

\para{Flexible Transport System.} Flexible transport systems make it easy to adjust the materials that the machines in a smart factory are supplied with. Most flexible transport systems transport materials with a team of agents~\cite{AMRs_in_manufacturing_operations}. These agents are generally autonomous mobile vehicles~\cite{AMRs_in_manufacturing_operations}. However, recent mag-lev based systems such as BOSCH's ctrlX Flow$^{\text{6D}}$~\cite{BOSCH_maglev}.



To embed a manufacturing procedure into a smart factory, the smart factory's operator needs to:
\begin{enumerate}
    \item assign each process in the manufacturing procedure to one or more machines in the smart factory.

    \item construct a transport plan that specifies how the smart factory's agents should carry parts between machines.
\end{enumerate}
The problem of embedding a manufacturing procedure in a smart factory is termed the Smart Factory Embedding (SFE) problem. A good embedding maximizes the smart factory's throughput, that rate that it makes finished products.

Most existing systems for coordinating agents in a smart factory assume processes have already been assigned to the smart factory's machines~\cite{sf_traffic_system, sf_dpn}. Assigning processes to 
machines and paths to agents separately, however, limits the throughput that these solvers can achieve.

One recent approach to the SFE problem, ACES~\cite{ACES}, jointly optimizes its two components. ACES models a smart factory as a grid of cells. Time is discretized. ACES uses a Mixed Integer Linear Program (MILP) to jointly assign processes to machines and find a transport plan to its agents. ACES generates a transport plan which loops after a certain number of timesteps, allowing it to be run continuously.

Unfortunately, ACES scales poorly. ACES's MILP has a binary variable which indicates if a given cell contains an agent carrying a given component on given timestep for every possible (cell, component, timestep) combination. A SFE instance may have hundreds of cells and tens of components. A transport plan may have tens of timesteps. As a result, when ACES is used to solve a large SFE instance, it may generate a MILP with 100,000s of these variables. A MILP with 100,000s of variables is often difficult to solve. Thus, to date, there is no solver which jointly optimizes both components of the SFE problem that works at scale. 

We address this hole by proposing the Traffic System based Cyclic Embedding Solver, \TSACES{}. \TSACES{} is based on the following observation: most smart factories coordinate agents using a traffic system, a network of roads.  \TSACES{} aggregates timesteps into epochs. It uses a MILP to construct a traffic system based embedding, an embedding which moves agents through a traffic system at the rate of one road per epoch. A traffic system often has several times fewer roads than cells. A traffic system based embedding often has several times fewer epochs than timesteps. \TSACES{}'s MILP thus often has dozens of times fewer variables than ACES's MILP, making it much easier to solve.


The throughput and runtime of \TSACES{}'s MILP are dependent on its hyperparameters, the number of epochs and the length of an epoch in the embedding that it generates. We introduce a novel,  principled search algorithm to find good values for these hyperparameters. \TSACES{} uses its traffic system based embedding to construct a transport plan generator. This generator  can generate timestep by timestep instructions for the smart factory's agents in real time, allowing it to run its manufacturing procedure indefinitely.



We analyze \TSACES{} and show that it is complete. Our evaluations show that \TSACES{} can scale to SFE instances based on real scenarios with more than 100 machines.

\section{Problem Formulation}

Our model of the SFE problem adapts the model given in~\cite{ACES} to smart factories that use traffic systems.

\subsection{Manufacturing Procedure}

\para{Token.} We model each assemblage, part or raw material produced or consumed during a manufacturing procedure as a token. The set of tokens associated with a manufacturing procedure is denoted $\allTk{} := \{\tk{1}, \tk{2}, \ldots\}$.

\para{Process.} A process $\p{i}$ is an atomic operation in a manufacturing procedure. Each process $\p{i}$ consumes a multiset of input tokens $\inTk{}(\p{i})$ and emits a multiset of output tokens $\outTk{}(\p{i})$. The number of copies of a token $\tk{j} \in \allTk{}$ that a process $\p{i}$ consumes and produces are denoted $\numIn{}(\p{i}, \tk{j})$ and $\numOut{}(\p{i}, \tk{j})$.

A process that does not consume any tokens is a source process. Source processes represent operations which retrieve raw materials. A process which does not produce any tokens is a sink process. Sink processes represent operations which remove finished products or waste. A manufacturing procedure's the set of processes is denoted $\allP{} := \{\p{1}, \p{2}, \ldots\}$. Exactly one of these processes $\outP{}(\allP{})$ must be an output process, a sink process which exports finished products.

\para{Manufacturing Procedure.} A manufacturing procedure $(\allTk{}, \allP{})$ is a set of tokens and a set of processes that consume and produce those tokens.

\para{Example.} Fig.~\ref{fig:example_problem}. (a) shows a manufacturing procedure for toy cars. It has the tokens $\allTk{} := \{\tk{p}, \tk{f}, \tk{a}, \tk{w}, \tk{c}\}$, the source processes $\p{1}$ and $\p{4}$, and the sink process $\p{6}$, which is also an output process. The tokens that a process consumes and produces are shown below it. Process $\p{5}$ consumes a chassis, 4 wheel and 2 axle tokens and produces a car token. 

\subsection{Smart Factory}

\para{Machines.} A smart factory has a set of machines $\allM{} := \{\m{1}, \m{2}, \ldots\}$. A machine $\m{i}$ can run a subset $\mProc{}(\m{i}) \subseteq \allP{}$ of a manufacturing procedure's processes. Time is discretized. A process $\p{j}$ takes a machine $\m{i}$ $\runtime{}(\m{i}, \p{j})$ timesteps to run. Machines have input and output buffers. To initiate a process $\p{j}$, a machine must consume the multiset of tokens $\inTk{}(\p{j})$ from its input buffer. When a machine finishes process $\p{j}$, it deposits the multiset of tokens $\outTk{}(\p{j})$ into its output buffer. Source and sink machines are special types of machines. Only source machines can run source processes. They represent bins of raw materials. Only sink machines can run sink processes. They represent chutes for waste and finished products.

\para{Layout.} We model the layout of a smart factory as a grid of square cells. Each cell has a set of entry cells and a set of exit cells, which must be vertically or horizontally adjacent. An agent can only enter a cell from its entry cells and leave a cell to its exit cells. Our model has three types of cells:

\begin{itemize}
    \item \textit{Road Cells.} A road cell has one entry and one exit cell. 

    \item \textit{Junction Cells.} A junction cell has at least one entry and exit cell. These cells must be road cells. 

    \item \textit{Non-Traversable Cells.} Agents may not enter non-traversable cells. Any cell which contains an obstacle like a machine chassis must be a non-traversable cell.
\end{itemize}

Let cell $\cl{i}$ be the $i$th road or junction cell in a layout. Let $\allCl{} := \{\cl{1}, \cl{2}, \ldots\}$ be the set of all road and junction cells in a layout. We represent the connections between the cells in a layout with a directed graph called a layout graph $\layoutG{} := (\allCl{}, \layoutE{})$. Each vertex in the layout graph is a road or junction cell. There is an arc $(\cl{i}, \cl{j}) \in \layoutE{}$ iff cell $\cl{i}$ is one of cell $\cl{j}$'s entry cells. Note that if cell $\cl{i}$ is one of cell $\cl{j}$'s entry cells, then cell $\cl{j}$ must be one of cell $\cl{i}$'s exit cells.

To be valid, a layout must have the following properties:
\begin{enumerate}
    \item Its layout graph must be strongly connected. If a layout graph is not strongly connected, agents may not be able to carry tokens between certain pairs of machines.

    \item It must contain at least one junction cell. In this paper, we exclude the trivial case where a factory's machines are connected by a single loop of road cells.
\end{enumerate}

Any machine's input buffer and output buffer is associated with an input cell and an output cell. An agent can only deposit tokens into an input buffer on its input cell and remove tokens from an output buffer on its output cell.


\begin{table}[t]
    \centering
    \begin{tabular}{l l l}
        \textbf{Machine Type}  &\textbf{Instances}   &\textbf{Supported Processes}\\
         Bin of Planks         &$\m{1}$                &$\p{1}$\\
         CNC Machine           &$\m{2}$, $\m{3}$, $\m{4}$  &$\p{2}$, $\p{3}$\\
         Assembler             &$\m{5}$                &$\p{5}$\\
         Bin of Axles          &$\m{6}$                &$\p{4}$\\
         Output Chute          &$\m{7}$                &$\p{6}$
    \end{tabular}
    \vspace{-1mm}
    \caption{The machines in the example smart factory.}
    \label{tab:example_machines}
    \vspace{-3mm}
\end{table}

\para{Example.} Fig.~\ref{fig:example_problem}. (b) depicts an example smart factory. Regular road cells are colored purple and junction cells green. Road cells which are also input and output cells are colored blue and yellow. An impassable cell is gray if it contains a machine's chassis and white if it empty. A machine's border with its input and output cells is indicated with a dotted line. For example, the cells $(5,1)$ and $(0,2)$ are machine $\m{5}$'s input and output cells. Table~\ref{tab:example_machines} describes this smart factory's machines. It has two source machines, $\m{1}$ and $\m{5}$, and one sink machine, $\m{6}$. 


\para{Traffic System.} We group the road and junction cells in a layout into roads and junctions.  A road $\road{i}$ is a path of $\rLen{}(\road{i})$ road cells which connects two junction cells. We denote the set of roads in a layout $\allRoads{} := \{\road{1}, \road{2}, \ldots\}$. A junction $\junction{i}$ is single junction cell. Each junction connects two or more roads. We denote the set of junctions in a layout $\allJunctions{} := \{\junction{1}, \junction{2}, \ldots\}$. A layout's roads and junctions collectively form a traffic system $(\allRoads{}, \allJunctions{})$. Fig.~\ref{fig:example_problem}. (c) shows the example smart factory's traffic system. It has 9 junctions $\allJunctions{} := \{\junction{1}, \ldots, \junction{9}\}$ and 12 roads $\allRoads{} := \{\road{1}, \ldots, \road{12}\}$. 

A junction $\junction{i}$ has a set of entry and exit roads $\entry{}(\junction{i})$ and $\exit{}(\junction{i})$. Any road which leads to and away from $\junction{i}$ is one of its entry and exit roads. For example, in the example smart factory, junction $\junction{8}$ has the entry roads $\entry{}(\junction{8}) = \{\road{11}\}$ and the exit roads $\exit{}(\junction{8}) = \{\road{9}, \road{11}\}$.



\subsection{Agents}

A team of $n$ agents $\allAg{} := \{\ag{1}, \ldots, \ag{n}\}$ carries tokens between machines. At the start of each timestep $t$, each agent $\ag{i} \in \allAg{}$ occupies a road or junction cell. We denote this cell $\agCl{}(\ag{i}, t) \in \allCl{}$. Each timestep, an agent $\ag{i} \in \allAg{}$ must either wait at its current cell or move to one of its current cell's exit cells. Two agents may not occupy the same cell or traverse the same edge in the layout graph on the same timestep.

Agents can carry a single token at a time. We term the token that an agent $\ag{i}$ is carrying its cargo. If an agent is not carrying a token, we say that it is carrying the null token $\tk{0}$. We denote the cargo that agent $\ag{i}$ is carrying on timestep $t$ $\agTk{}(\ag{i}, t) \in \allTk{} \cup \{\tk{0}\}$. The state $(\agCl{}(\ag{i}, t), \agTk{}(\ag{i}, t))$ of agent $\ag{i}$ on timestep $t$ is its location and cargo.

If an agent carrying a non-null token is on a machine's input cell at the end of a timestep $t$, it may deposit its token into the machine's input buffer. If an agent carrying a null token is on a machine's output cell at the end of a timestep $t$, it may pick up a token from the machine's output buffer.


\subsection{Cell Based Embedding}

A cell based embedding specifies how a manufacturing procedure is implemented by a smart factory. It is a 3-tuple  $(\assignM{}, \rateM{}, \tpGen{})$ with the following elements:

\para{Assignment Matrix.} An assignment matrix $\assignM{}$ is a $
|\allM{}| \times |\allP{}|$ matrix. The field $\assign{}(\m{i}, \p{j})$ contains a binary variable which indicates if machine $\m{i}$ has been assigned process $\p{j}$. A machine may only be assigned one process.

\para{Rate Matrix.} A rate matrix $\rate{}$ is also a $|\allM{}| \times |\allP{}|$ matrix. The field $\rate{}(\m{i}, \p{j})$ indicates the rate that machine $\m{i}$ runs process $\p{j}$. A machine $\m{i}$ can only run process $\p{j}$ at a non-zero rate if it is assigned process $\p{j}$. 


\para{Cell Based Transport Plan.} A cell based transport plan specifies how agents carry tokens through a smart factory. A smart factory often has to run a manufacturing procedure for an indefinite length of time. We would thus like to construct a transport plan generator $\tpGen{}$, a function which takes a smart factory's state $\sfState{}(t)$ at any timestep $t$ and computes its state $\sfState{}(t+1)$ at timestep $t+1$. As long as our generator can run in real time, we can run a manufacturing procedure endlessly.

The state $\sfState{}(t)$ of a smart factory at timestep $t$ is a 4-tuple $(\bInM{}(t), \bOutM{}(t), \agClM{}(t), \agTkM{}(t))$ with the following components:

\para{Buffer Contents Vectors.} The input and output buffer contents vectors $\bInM{}(t)$ and $\bOutM{}(t)$ are length $|\allM{}|$ vectors. The fields $\bIn{}(\m{i}, t)$ and $\bOut{}(\m{i}, t)$ specify the contents of a machine $\m{i}$'s input and output buffers on timestep $t$. If machine $\m{i}$ does not have an input buffer, $\bIn{}(\m{i}, t) = \emptyset$. If it does not have an output buffer, $\bOut{}(\m{i}, t) = \emptyset$.

\para{Agent State Vectors.} The position and cargo vector $\agClM{}(t)$ and $\agTkM{}(t)$ are length $|\allAg{}|$ vectors. The fields $\agClM{}(\ag{i}, t)$ and $\agTkM{}(\ag{i}, t)$ specify agent $\ag{i}$'s position and cargo on timestep $t$.

\subsection{Smart Factory Embedding Problem}
The throughput $\throughput{}(\assignM{}, \rateM{}, \tpGen{}) := \sum_{\m{i} \in \allM{}} \rate{}(\m{i},\outP{}(\allP{}))$  of an embedding $(\assignM{}, \rateM{}, \tpGen{})$ is the total rate at which its machines run its output process. In the smart factory embedding problem, we are given a manufacturing procedure $(\allTk{}, \allP{})$, and a smart factory $(\allM{}, \allCl{}, \allAg{}, \allJunctions{}, \allRoads{})$ and asked to find a maximum throughput embedding. We term the tuple $(\allTk{}, \allP{}, \allM{}, \allCl{}, \allAg{}, \allJunctions{}, \allRoads{})$ a SFE instance and denote it $\inst{}$.

\section{Related Work}

The Multi-Agent Path Finding (MAPF) problem is the problem of moving a team of agents  from start locations to goal locations without collisions. MAPF is an key part of SFE. MAPF has been solved via a range of methods such as SAT solving~\cite{MAPF_SAT} and answer set programming~\cite{ASP_LMAPF_offline}. 

Many other optimization problems involve MAPF. One problem which incorporates MAPF and is related to the SFE problem is the Collective Construction (CC) problem. In the CC problem, a team of agents is tasked with arranging building blocks into a structure. The agents are the same size as the blocks, forcing them to climb the structure to position the block. The CC problem is formalized as a combinatorial optimization problem and solved for a single agent using dynamic programming~\cite{single_agent_CCP}. This approach is generalized to multiple agents in~\cite{dp_CCP}. Empirically, however, the solutions that it generates do not achieve much parallelism. The reinforcement learning approach proposed in~\cite{rl_CCP} finds solutions with more parallelism. The CC problem is solved optimally using both constraint set programming and a MILP in~\cite{CSP_MILP_CCP}.

Several systems for coordinating agents in a smart factory have been proposed. In~\cite{sf_traffic_system}, the authors use a traffic system to traffic system to generate paths for agents in a manufacturing facility. In~\cite{sf_dpn}, the authors assign tasks to agents and plan their paths using a distributed petri net. Neither system, however, assigns processes to machines. ACES~\cite{ACES} solves the SFE problem using a MILP. It is unclear, however, whether it can scale to SFE instances with more $\sim 25$ machines.

\section{Overview}

 \TSACES{} solves the SFE by constructing a traffic system based embedding and then converting it into a cell based embedding. In this section, we define a traffic system based embedding and then describe \TSACES{}'s workflow.


\para{Traffic System Based Embedding.} A traffic system based embedding is based on roads, not cells. It aggregates timesteps into epochs. It moves agents one road per epoch. At the start of an epoch, each agent is in a queue at the head of a road. During an epoch, each agent passes through the junction $\junction{i}$ at the end of its road and joins a queue on one of $\junction{i}$'s exit roads. The number of timesteps in an epoch is denoted $\epochLen{}$.

A manufacturing procedure often has to be run for an indefinite length of time. We therefore construct an embedding with a cyclic transport plan~\cite{CCPP}, a transport plan that can be looped repeatedly. A cyclic transport plan is associated with a number of epochs $\numEpochs{}$. It loops after $\numEpochs{}$ epochs.

A traffic system based embedding can be written as an 6-tuple: $(\assignM{}, \rateM{}, \bInM{}, \bOutM{}, \pkM{}, \dpM{})$. Its assignment and rate matrices $\assignM{}$ and $\rateM{}$ are written using the same matrices as a cell based embedding. Its transport plan is expressed as a 4 tuple $(\inBM{}, \outBM{}, \pkM{}, \dpM{})$ with the following new components:

\para{Inbound and Outbound Traffic Tensors.} The inbound and outbound traffic tensors $\inBM{}$ and $\outBM{}$ are $|\allRoads{}| \times \numEpochs{}+1 \times |\allTk{}|$ tensors. The fields $\inB{}(\road{i}, \T{}, \tk{j})$ and $\outB{}(\road{i}, \T{}, \tk{j})$ specify the number of agents carrying token $\tk{j}$ that enter and leave road $\road{i}$ during epoch $\T{}$.

\para{Pick Up and Deposit Tensors.} The pickup and deposit tensors $\pkM{}$ and $\dpM{}$ are $|\allM{}| \times \numEpochs{}+1 \times |\allAg{}|$ tensors. Let $\road{out}$ and $\road{in}$ be the roads that contain machine $\m{i}$'s output and input cells. The field $\pk{}(\m{i}, \T{}, \tk{j})$ specifies the number of agents entering road $\road{out}$ during epoch $\T{}$ that pick up a copy of token $\tk{j}$ from $\m{i}$'s output buffer. The field $\dpF{}(\m{i}, \T{}, \tk{j})$ specifies the number of agents entering $\road{in}$ during epoch $\T{}$ that pick up a copy of $\tk{j}$ from $\m{i}$'s input buffer.

Since our embedding loops after $\numEpochs{}$ epochs, the values of its tensors for the epoch $\T{} = 0$ and $\T{} = \numEpochs{}$ are the same.

\begin{figure}[t]
    \centering
    \includegraphics[width=0.85\linewidth, trim={0 0 0 0},clip]{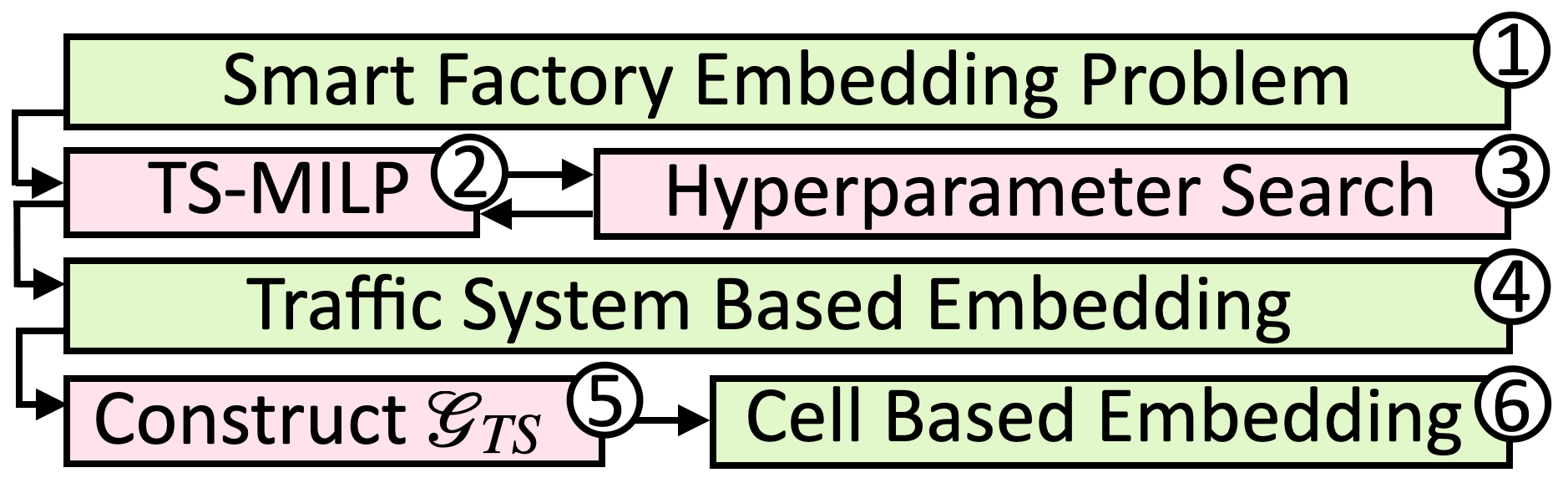}
    \vspace{-3mm}
    \caption{\TSACES{}'s workflow.}
    \label{fig:tsaces_arch}
\end{figure}




\para{\TSACES{}'s Workflow.} The workflow of \TSACES{}  is shown in Fig.~\ref{fig:tsaces_arch}. \TSACES{} uses a MILP termed the \TSMILP{} $\circled{2}$ to generate a traffic system based embedding \circled{4} for a SFE instance \circled{1}. We describe the \TSMILP{} in Section~\ref{sec:gen_traffic_system_based_embedding}.  The \TSMILP{}'s solution quality and runtime depend heavily on its hyperparameters, the number of epochs $\numEpochs{}$ and the length of an epoch $\epochLen{}$ in the embedding that it is asked to find. \TSACES{} uses a hyper-parameter search algorithm~\circled{3} to find a pair of hyperparameters which  produce a good solution. We describe this algorithm in Section~\ref{sec:hyperparameter_search}. 


Once \TSACES{} has finished searching for a traffic system based transport plan, it uses that it to construct a cell based transport plan generator $\TSTGEN{}$. $\TSTGEN{}$ implements \TSACES{}'s  transport plan. For example, $\TSTGEN{}$ moves $\inB{}(\road{i}, \T{}, \tk{j})$ agents carrying token $\tk{j}$ into road $\road{i}$ on any epoch $\T{}' = \T{} \% \numEpochs{}$. It takes $O(|\allAg{}|)$ time for $\TSTGEN{}$ to compute the next state in its transport plan. We describe $\TSTGEN{}$ in Section~
\ref{sec:traffic_system_based_embedding_to_cell_based_embedding}.



\section{Generating a Traffic System Based Embedding}
\label{sec:gen_traffic_system_based_embedding}

In this section, we specify the \TSMILP{}, the MILP used to construct a traffic system based embedding. 


\para{Objective.} The \TSMILP{} maximizes the throughput of its embedding that it generates:
%
\begin{equation*}
    \max \sum_{\m{i} \in \allM{}} \rate{}(\m{i},\outP{}(\allP{}))
\end{equation*}

The \TSMILP{} has 5 categories of constraints. 


\para{Machine Configuration Constraints.} These constraints specify how the machines in a smart factory can be configured.

\para{Constraint 1.} A machine can only be assigned one process.
\begin{equation*}
\forall\ \m{i} \in \allM{}, \sum_{\p{j} \in \allP{}} \assign{}(\m{i}, \p{j}) \leq 1.
\end{equation*}

\para{Constraint 2.} A machine must be able to run its process.
\begin{equation*}
    \forall\ \m{i} \in \allM{},\ \forall\ \p{j} \in \allP{} \setminus \mProc{}(\m{i}),\  \assign(\m{i},\p{j}) =0.
\end{equation*}

\para{Constraint 3.} A machine $\m{i} \in \allM{}$ can only run the process $\p{j} \in \mProc{}(\m{i})$ once every $\runtime{}(\m{i}, \p{j})$ timesteps.
\begin{equation*}
    \forall \m{i} \in \allM{},\ \forall \p{j} \in \mProc{}(\m{i}),  \rate(\m{i}, \p{j}) \leq \runtime{}(\m{i},\p{j})^{-1}
\end{equation*}

\para{Constraint 4.} A machine $\m{i} \in \allM{}$ can only run the process $\p{j} \in \allP{}$ at a non-zero rate if it is assigned $\p{j}$.
\begin{equation*}
    \forall\ \m{i} \in \allM{},\ \forall\ \p{j} \in \allP{},\ \rate{}(\m{i},\p{j}) - \assign{}(\m{i},\p{j}) \leq 0.
\end{equation*}

\para{Buffer Conservation Constraints.} These constraints ensure that tokens don't appear in and disappear from buffers.


\para{Constraint 5.} Each cycle, the number of copies of a token $\tk{j}$ that a machine $\m{i}$ produces and the number of copies of $\tk{j}$ that agents pick up from its output buffer must be the same.
\begin{align*}
    &\forall\ \m{i} \in \allM{} - \sinkM{},\ \forall\ \tk{j} \in \allTk{}, \
    \sum_{\T{}=0}^{\numEpochs{}-1} \pk{}(\m{i}, \T{}, \tk{j}) =\\ %
    &\sum_{\p{k} \in \allP{}} \rate(\m{i},\p{k}) \cdot \numOut(\p{k}, \tk{j}) \cdot \numEpochs{} \cdot \epochLen{}.%
\end{align*}

    


\para{Constraint 6.} Each cycle, the number of copies of a token $\tk{j}$ that a machine $\m{i}$ consumes and the number of copies of $\tk{j}$ that agents deposit in its input buffer must be the same.
\begin{align*}
    &\forall\ \m{i} \in \allM{} - \srcM{},\ \forall\ \tk{j} \in \allTk{}, \
    \sum_{\T{}=0}^{\numEpochs{}-1} \dpF{}(\m{i}, \T{}, \tk{j}) =\\ %
    &\sum_{\p{k} \in \allP{}} \rate(\m{i},\p{k}) \cdot \numIn(\p{k}, \tk{j}) \cdot \numEpochs{} \cdot \epochLen{}.%
\end{align*}



\para{Traffic System Conservation Constraints.} These constraints ensure that agents and tokens don't appear or disappear. Let $\inputOn{}(\road{i}) \subseteq \allM{}$ and $\outputOn{}(\road{i}) \subseteq \allM{}$ be the sets of machines whose input and output cells are on road $\road{i}$.

\para{Constraint 7.} The number of agents carrying a non-null token $\tk{j} \in \allTk{}$ leaving the road $\road{i}$ during epoch $\T{} + 1\ \%\ \numEpochs{}$ is the number of agents carrying token $\tk{j}$ entering road $\road{i}$ during epoch $\T{}$, \textit{less} the number of these agents which deposit their copy of $\tk{j}$ before leaving $\road{i}$, \textit{plus} the number of agents which enter $\road{i}$ during epoch $\T{}$ that pick up a copy of $\tk{j}$.
\begin{align*}
\forall\ &\road{i} \in \allRoads{},\ \forall\ \tk{j} \in \allTk{},\ \forall\ \T{} \in [0..\numEpochs{}-1],\\ %
&\outB{}(\road{i}, \T{} + 1\ \%\ \numEpochs{}, \tk{j}) =\\ %
&\inB{}(\road{i}, T, \tk{j}) 
- \sum_{\mathclap{\m{k} \in \inputOn{}(\road{i})}} \dpF{}(\m{k}, T, \tk{j}) %
+ \sum_{\mathclap{\m{k} \in \outputOn{}(\road{i})}} \pk{}(\m{k}, T, \tk{j}).
\end{align*}

\para{Constraint 8.} Similarly, the number of agents carrying a null token  leaving the road $\road{i}$ during epoch $\T{} + 1\ \%\ \numEpochs{}$ is the number of agents carrying token $\tk{0}$ entering $\road{i}$ during epoch $\T{}$, \textit{less} the number of these agents which pick up a token before leaving $\road{i}$, \textit{plus} the number of agents entering $\road{i}$ during epoch $\T{}$ which deposit a token.
\begin{align*}
&\forall\ \road{i} \in \allRoads{},\ \forall\ \T{} \in [0..\numEpochs{}-1],\  %
\outB{}(\road{i}, \T{} + 1\ \%\ \numEpochs{}, \tk{0}) =\\ %
&\inB{}(\road{i}, T, \tk{0}) 
- \sum_{\mathclap{\m{k},\tk{j} \in \outputOn{}(\road{i}) \times   \allTk{} \quad}} \pk{}(\m{k}, T, \tk{j}) %
+ \sum_{\mathclap{\quad \m{k},\tk{j} \in \inputOn{}(\road{i}) \times \allTk{} }} \dpF{}(\m{k}, T, \tk{j}).
\end{align*}

\para{Constraint 9.} Every agent which enters a junction $\junction{i} \in \allJunctions{}$ during epoch $\T{}$ must leave junction $\junction{i}$ during epoch $\T{}$.
\begin{align*}
\forall\ \junction{i} \in\ &\allJunctions{},\ \forall\ \T{} \in [0..\numEpochs{}-1],\ \forall\ \tk{j} \in \allTk{} \cup \{\nullTk\},\\ %
 &\sum_{\mathclap{\road{k} \in \exit{}(\junction{i})}}  \inB{}(\road{k}, \T{}, \tk{j}) %
= \sum_{\mathclap{\road{k} \in \entry{}(\junction{i})}} \outB{}(\road{k}, \T{}, \tk{j}).
\end{align*}

\para{Agent Behavior and Team Size Constraints.} After an agent changes its cargo, it must move to a new road before changing its cargo again. The \TSMILP{} does not know how a road's input and output cells are ordered. If a road $\road{i}$'s output cells come after its input cells, an agent on road $\road{i}$ will not be able to deposit and then pick up a token (and vice versa). Consequently, the \TSMILP{} can only rely on agents being able to change their token once per road.


\para{Constraint 10.} The number of agents that deposit a copy of token $\tk{j}$ on road $\road{i}$ during epoch $\T{}$ must be less than the number of agents that enter $\road{i}$ carrying $\tk{j}$ during epoch $\T{}$.
\begin{align*}
\forall\ \road{i} \in \allRoads{},\ \forall\ \tk{j} \in \allTk{},\ \forall&\ \T{} \in [0..\numEpochs{}-1],\\ %
&\inB{}(\road{i}, \T{}, \tk{j}) %
\leq \dpF{}(\road{i}, \T{}, \tk{j}).%
\end{align*}

\para{Constraint 11.} The number of agents that pick up token $\tk{j}$ on road $\road{i}$ during epoch $\T{}$ must be less than the number of agents that enter $\road{i}$ carrying the null token during epoch $\T{}$.
\begin{align*}
\forall\ \road{i} \in \allRoads{},\ \forall\ \tk{j} \in \allTk{},\ \forall&\ \T{} \in [0..\numEpochs{}-1],\\ %
&\inB{}(\road{i}, \T{}, \tk{j}) %
\leq \dpF{}(\road{i}, \T{}, \tk{j}).%
\end{align*}

\para{Constraint 12.} The \TSMILP{} cannot produce a solution which uses more agents than the $|\allAg{}|$ agents available. 
\begin{align*}
    \sum_{\road{i} \in \allRoads{}} \sum_{\tk{j} \in \allTk{} \cup \{\tk{0}\}} \outB{}(\road{i}, 0, \tk{j}) \leq |\allAg{}|.
\end{align*}

\para{Road Capacity Constraints.} These constraints limit the number of agents which pass through each road every epoch. Let $\totalInB{}(\road{i}, \T{}) {:=} \sum_{\tk{j} \in \allTk{} \cup \{\tk{0}\}} \inB{}(\road{i}, \T{}, \tk{j})$ and $\totalOutB{}(\road{i}, \T{}) {:=} \sum_{\tk{j} \in \allTk{} \cup \{\tk{0}\}} \outB{}(\road{i}, \T{}, \tk{j})$ be the number of agents that enter and leave road $\road{i}$ on epoch $\T{}$.


\para{Constraint 13.} Our ILP cannot specify when a road $\road{i}$'s outbound agents will leave or its inbound agents arrive during any given epoch. Its inbound agents may all arrive before any of its outbound agents leave. Every agent inbound to and outbound from a road $\road{i}$ during any epoch $\T{}$ must therefore be able to fit on road $\road{i}$ at the same time.
\begin{align*}
\forall\ \road{i} &\in \allRoads{}, \forall\ \T{} \in  [0..\numEpochs{}-1],\\ %
&\totalInB{}(\road{i}, T, \tk{j}) + \totalOutB{}(\road{i}, T, \tk{j})
\leq \rLen{}(\road{i}).
\end{align*}

\para{Constraint 14.} Each epoch, every agent must move from the queue at the end of one road to the queue at the end of a different road. If there are a lot of agents on a junction's entry roads, it will take them a long time to reach these queues on its exit roads. We ensure that this process doesn't take longer than one epoch by limiting the number of agents which can be on a junction's entry roads at the start of any epoch.

\begin{theorem}
\label{thm:time_to_reach_q}
An agent in a queue on one of a junction $\junction{i} \in \allJunctions{}$'s entry roads at the start of epoch $\T{}$ takes at most:
\begin{equation*}
    \sum_{\mathclap{\ \quad \quad \quad \road{j} \in \entry{}(\junction{i})}} \totalOutB{}(\road{j}, \T{}) +\ 
    \max_{\raisebox{-0.7ex}{$\scriptstyle \mathclap{\road{j} \in \exit{}(\junction{i})}$}}\ 
    [\rLen{}(\road{j}) - \totalInB{}(\road{j}, \T{})] + 1
\end{equation*}
timesteps to reach a queue on any one of $\junction{i}$'s exit roads.

\para{Proof.} Let $\ag{k}$ be the last agent to enter junction $\junction{i}$ on epoch $\T{}$. Agent $\ag{k}$ has to wait $\sum_{\road{j} \in \entry{}(\junction{i})} \totalOutB{}(\road{j}, \T{}) - 1$ timesteps for the other agents on $\junction{i}$'s entry roads to pass through $\junction{i}$. Entering $\junction{i}$ takes $\ag{k}$ an additional timestep. 

Agent $\ag{k}$ then has to reach the queue at the end of one of $\junction{i}$'s exit roads. Let $\road{j} \in \exit{}(\junction{i})$ be the exit road whose queue takes $\ag{k}$ the longest to reach. Agent $\ag{k}$ will be the last agent on road $\road{j}$ to reach its queue. Thus, when $\ag{k}$ reaches the queue, it will be $\totalInB{}(\road{i}, \T{}) - 1$ agents long. Agents on a road move one cell per timestep until the reach its queue. It therefore takes $\ag{k}$ $\rLen{}(\road{j}) - \totalInB{}(\road{j}, \T{})] + 1$ timesteps to reach road $\road{j}$'s queue. $\square$
\end{theorem}

By Theorem~\ref{thm:time_to_reach_q}, we can ensure that all agents reach the queue at the end of their road with the constraint:
\begin{align*}
\forall\ \junction{i}& \in \allJunctions{},\ \forall\ \road{j}\in \exit{}(\junction{i}),\ \forall\ \T{} \in [0..\numEpochs{}-1],\
\epochLen{} \geq\\
&\sum_{\mathclap{\road{k} \in \entry{}(\junction{i})}} \totalOutB{}(\road{k}, \T{})
+ \rLen{}(\road{j}) - \totalInB{}(\road{j}, \T{})
+ 1.
\end{align*}

\section{Hyperparameter Search}
\label{sec:hyperparameter_search}

In this section, we introduce an algorithm which finds good values for the number of epochs $\numEpochs{}$ and the length of an epoch $\epochLen{}$ in the embedding that the \TSMILP{} constructs.

\begin{algorithm}[t]
\small
\caption{\tsPlanner{}{}($\inst{}, \timer{}$)}
\label{alg:tsPlanner}
\begin{algorithmic}[1]
%
\State $\sol{}^*, \numEpochs{}^*, \epochLen{}^* \gets \nullVal{}, \nullVal{}, \nullVal{}$
\State $\numEpochs{} \gets 1$ 
\label{ln:make_num_epochs}

\While{$\timer{}$ has not expired}
\label{ln:checkTimer}
  \State $\sol{}, \epochLen{} \gets  \planForNumEpochs{}(\inst{}, \timer{})$
  \label{ln:plan_for_num_epochs}
  \If{$\sol{} \neq \nullVal{} \wedge \throughput{}(\sol{}) > \throughput{}(\sol{}^*)$}
  \label{ln:check_solution_throughput}
    \State $\sol{}^*, \numEpochs{}^*, \epochLen{}^* \gets \sol{}, \numEpochs{}, \epochLen{}$
    \label{ln:update_best_solution}
  \EndIf
  \State $\numEpochs{} \gets \numEpochs{} + 1$
  \label{ln:increment_num_epochs}
\EndWhile
\State \Return $\sol{}^*, \numEpochs{}^*, \epochLen{}^*$
\label{ln:return_sol}
\end{algorithmic}
\end{algorithm}

\para{Number of Epochs.} The space of solutions to the \TSMILP{} with $\numEpochs{}$ epochs can be very different to the space of solutions to the \TSMILP{} with $\numEpochs{} + 1$ epochs. We would therefore like our solver to try to solve the \TSMILP{} with as many different values of $\numEpochs{}$ as possible. We accomplish this using the function $\tsPlanner{}$. $\tsPlanner{}$ is shown in Algorithm~\ref{alg:tsPlanner}.

\para{\tsPlanner{}.} Let $\timer{}$ be a timer which tracks the time allocated to $\tsPlanner{}$. Let $\planForNumEpochs{}(\inst{},$ $\numEpochs{}, \timer{})$ be a function which returns an optimized solution $\sol{}$ to the \TSMILP{} for the SFE instance $\inst{}$ with $\numEpochs{}$ epochs and its epoch length $\epochLen{}$. If the timer expires while this function is running, it returns its current best solution if one exists and $(\nullVal{}, \nullVal{})$ otherwise.

The number of variables in the \TSMILP{} increases with the number of epochs in its transport plan. Solving the \TSMILP{} with a small number of epochs is thus usually faster than solving it with a large number of epochs. $\tsPlanner{}$ therefore begins by passing the \TSMILP{} a small value of $\numEpochs{}$. It then solves the \TSMILP{} with progressively larger values of $\numEpochs{}$ until the $\timer{}$ runs out, whereupon it returns $\sol{}^*$, the highest throughput solution that it has found. 


\begin{algorithm}[t]
\small
\caption{\planForNumEpochs{}$(\inst{}, \numEpochs{}, \timer{})$}
\label{alg:planForNumEpochs}
\begin{algorithmic}[1]
\State $\sol{}^*, \epochLen{}^* \gets \nullVal{}, \nullVal{}$
\State $\epochLen{} \gets \max_{\road{i} \in \allRoads{}} \rLen{}(\road{i}) + \incrEpochLen{}$
\label{ln:initialize_epoch_len}
%
\State $\runsSinceBestSol{} \gets 0$
\While{$\timer{}$ has not expired and $\runsSinceBestSol{} < \maxFailureNum{}$}
\State $\sol{} \gets \solveTSILP{}(\inst{}, \numEpochs{}, \epochLen{}, \timer{})$
\label{ln:solve_ILP}
\If{$\sol{} \neq \nullVal{} \wedge \throughput{}(\sol{}) > \throughput{}(\sol{}^*)$}
\State $\sol{}^*, \epochLen{}^* \gets \sol{}, \epochLen{}$
\State $\runsSinceBestSol{} \gets 0$
\Else{}
\State $\runsSinceBestSol{} \gets \runsSinceBestSol{} + 1$
\EndIf 
\State $\epochLen{} \gets \epochLen{} + \incrEpochLen{}$
\label{ln:incr_epoch_len}
\EndWhile
\end{algorithmic}
\end{algorithm}

\para{Epoch Length.} Most epoch lengths do not produce good solutions. If our solver's epoch length is too small, only a couple of agents can pass through a junction every epoch. If its epoch length is too large, agents spend most of their time waiting in a queue for the next epoch to start. We pick epoch lengths heuristically using the function \planForNumEpochs{}, shown in Algorithm~\ref{alg:planForNumEpochs}.




\para{\planForNumEpochs{}.} Let $\solveTSILP{}(\inst{},\numEpochs{}, \epochLen{},$ $ \timer{})$ be a function which uses the \TSMILP{} to solve the TS-SFE instance $\inst{}$ with $\numEpochs{}$ epochs and the epoch length $\epochLen{}$. If the $\timer{}$ expires before it has finished, it returns its current best solution if one exists and $\nullVal{}$ otherwise.

\planForNumEpochs{} starts by solving the \TSMILP{} with an epoch length $\epochLen{}$ intended to be smaller than optimal. It then solves the \TSMILP{} for progressively larger values of $\epochLen{}$. When $\epochLen{}$ grows larger than optimal, the throughput of \solveTSILP{}'s solutions will stop improving. Let $\sol{}^*$ be \solveTSILP{}'s best solution. If $\maxFailureNum{}$ consecutive epoch length increases pass without $\sol{}^*$ improving, the function terminates.

We increase \solveTSILP{}'s epoch length by $\incrEpochLen{}$ every run. Small values of $\incrEpochLen{}$ improve \planForNumEpochs{}'s solution quality but increase its runtime, large values do the opposite.

 \TSMILP{} needs an epoch length of $\max_{\road{i} \in \allRoads{}} \rLen{}(\road{i}) + 1$ or more to allow agents to traverse every road in the traffic system. $\planForNumEpochs{}$ initially runs $\solveTSILP{}$ with an epoch length $\incrEpochLen{}$ timesteps longer than this minimum. 


\section{Generating a Cell Based Embedding}
\label{sec:traffic_system_based_embedding_to_cell_based_embedding}

\begin{algorithm}[t]
\small
\caption{$\TSTGEN{}{}(\sfState{}(t), t)$}
\label{alg:TS_ACES}
\begin{algorithmic}[1]
%
\State \textbf{global} $(\allTk{}, \allP{}, \allM{}, \allAg{}, \allJunctions{}, \allRoads{})$
\label{ln:load_SFE_instance}
\State \textbf{global} $(\assignM{}, \rateM{}, \inBM{}, \outBM{}, \pkM{}, \dpM{}), \numEpochs{}, \epochLen{}$
\label{ln:load_traffic_system_based_embedding}
%
\State \textbf{global} $\canChgTk{}, \arrivalT{}$
\label{ln:load_canChngTk_and_arrivalT}
\State \textbf{global} $\inBM{}', \outBM{}', \pkM{}', \dpM{}'$ 
\label{ln:load_tensors}
%
\State \textbf{global} $(\bInM{}(t), \bOutM{}(t), \agClM{}(t), \agTkM{}(t)) \gets \sfState{}(t)$
\label{ln:load_sfState}
\State \textbf{global} $\atM{}(t) \gets \genOccupancyVec{}(\agClM{}(t))$
\label{ln:genOccupancyVec}
%
%
\State \textbf{initialize} $\bInM{}(t+1), \bOutM{}(t+1), \agTkM{}(t+1), \atM{}(t+1)$
\label{ln:init_new_sfState}
%
\State $\T{} \gets \lfloor t / \epochLen{} \rfloor$
%
%
\If{$t\ \%\ \epochLen{} = 0$}
\label{ln:is_epoch_start}
  \State $\adjustStateForNewEpoch{}()$
  \label{ln:adjustStateForNewEpoch}
\EndIf
%
\For{$\junction{h} \in \allJunctions{}$}
\label{ln:iterate_over_junctions}
  \State $\ag{i} \gets{} \agentAt{}(\cell{}(\junction{j}), t-1)$
  \If{$\ag{i} \neq \nullVal{}$}
    \State $\moveAgentOnJunction{}(\junction{h}, \ag{i}, t, \T{})$
    \label{ln:move_agent_on_junction}
  \EndIf
\EndFor
\For{$\road{h} \in \allRoads{}$}
\label{ln:iterate_over_roads}
  \State $\moveAgentsOnRoad{}(\road{h}, t, \T{})$
  \label{ln:move_agents_on_road}
\EndFor
\State $\depositToken{}(\allM{}, t, \T{})$
\label{ln:deposit_token}
\State $\pickupToken{}(\allM{}, t)$
\label{ln:pickup_token}
\State $\keepToken{}(t)$
\label{ln:keep_token}
\State $\agClM{}(t+1) \gets \genPositionVec{}(\atM{}(t+1))$
\label{ln:genPositionVec}
%
%
\State \Return $(\bInM{}(t+1), \bOutM{}(t+1), \agClM{}(t+1), \agTkM{}(t+1))$
%
\Function{$\moveAgentOnJunction{}$}{$\junction{h}, \ag{i}, t, \T{}$}
\label{ln:move_agent_on_junction_begin}
\State $\tk{j} \gets \agTk{}(\ag{i}, t)$
\label{ln:maoj_get_ag_token}
\State $\road{k} \gets \rndEle{}(\{\road{l} \in \exit{}(\junction{h}) : \inB{}(\road{l}, \T{}, \tk{j}) > 0\})$
\label{ln:maoj_identify_next_road}
\State $\agentAt{}(\tail{}(\road{k}), t+1) \gets \ag{i}$
\label{ln:move_agent_to_exit_road}
\State $\inB{}'(\road{k}, \T{}, \tk{j}) \gets \inB'{}(\road{k}, \T{}, \tk{j}) - 1$
\label{ln:update_inB}
\State $\canChgTk{} \gets \canChgTk{} \cup \{\ag{i}\}$
\label{ln:update_canChgTk}
\State $\arrivalT{}(\ag{i}) \gets \T{}$
\label{ln:update_arrivalT}
\EndFunction
\end{algorithmic}
\end{algorithm}

\TSACES{} uses the traffic system based embedding generated by the \TSMILP{} to construct the cell based transport plan generator $\TSTGEN{}$. $\TSTGEN{}$ takes the smart factory's state $\sfState{}(t)$ at timestep $t$ and generates its state $\sfState{}(t+1)$ at timestep $t+1$. $\TSTGEN{}$ is shown in Algorithm~\ref{alg:TS_ACES}. $\TSTGEN{}$ begins by loading the following global state variables:
\begin{enumerate}
    \item The SFE instance that \TSACES{} is solving (Line~\ref{ln:load_SFE_instance}) and its traffic system based embedding (Line~\ref{ln:load_traffic_system_based_embedding}).

    \item The set $\canChgTk{}$, which contains every agent which is allowed to change tokens on timestep $t$ (Line~\ref{ln:load_canChngTk_and_arrivalT}).

    \item The length $|\allAg{}|$ entry epoch vector $\arrivalT{}$. The field $\arrivalT{}(\ag{i})$ stores the epoch during which agent $\ag{i}$ entered its current road (Line~\ref{ln:load_canChngTk_and_arrivalT}).

    \item The maps $\inBM{}', \outBM{}', \pkM{}'$ and $ \dpM{}'$ (Line~\ref{ln:load_tensors}). The fields $\inB{}'(\road{i}, \T{}, \tk{j})$ and $\outB{}'(\road{i}, \T{}, \tk{j})$ store the number of additional agents carrying token $\tk{j}$ that $\TSTGEN{}$ need to move into and out of road $\road{k}$ during epoch $\T{}$. The fields $\pk{}'(\m{h}, \T{}, \tk{j})$ and $\dpF{}'(\m{h}, \T{}, \tk{j})$ store the number of additional copies of $\tk{j}$ that need to be picked up from and deposited into $\m{h}$'s buffers by agents that entered their current road during epoch $\T{}$. 
\end{enumerate}
$\TSTGEN{}$ then loads the smart factory's state (Line~\ref{ln:load_sfState}).

\para{\genOccupancyVec{}.} $\TSTGEN{}$ stores the positions of its agents in an occupancy vector $\atM{}(t)$, a length $|\allCl{}|$ vector (Line~\ref{ln:genOccupancyVec}). The field $\at{}(\cl{j}, t)$ stores the agent on cell $\cl{j}$ on timestep $t$. If no agent is on $\cl{j}$ on timestep $t$, $\at{}(\cl{j}, t) = \nullVal{}$.

$\TSTGEN{}$ then initializes the vectors storing the smart factory's state at timestep $t+1$ (Line~\ref{ln:init_new_sfState}). Their fields are set to $\nullVal{}$.

\para{\adjustStateForNewEpoch{}.} If timestep $t$ is the start of a new epoch $\T{}$ (Line~\ref{ln:is_epoch_start}), $\TSTGEN{}$ adjusts its state variables (Line~\ref{ln:adjustStateForNewEpoch}) with the function  \adjustStateForNewEpoch{}. It adds a field $\inB{}'(\road{i}, \T{}, \tk{j})$ to $\inB{}'$ for every (road, token) combination. This field is set to $\inB{}(\road{i}, \T{} \% \numEpochs{}, \tk{j})$ since no agents carrying token $\tk{j}$ have moved into road $\road{i}$ on epoch $\T{}$. It removes each field from the epoch $\T{}-2$. It performs analogous operations on the maps $\outBM{}'$, $\pkM{}'$ and $\dpM{}$.

\para{\moveAgentOnJunction{}.} Next, $\TSTGEN{}$ moves each agent $\ag{i}$ on a junction $\junction{h}$'s cell $\cell{}(\junction{h})$ into one of its exit roads with the function \moveAgentOnJunction{} (Lines~\ref{ln:iterate_over_junctions}-\ref{ln:move_agent_on_junction}, \ref{ln:move_agent_on_junction_begin}-\ref{ln:update_arrivalT}). Let the token $\tk{j}$ be agent $\ag{i}$'s cargo (Line~\ref{ln:maoj_get_ag_token}). $\TSTGEN{}$ identifies a road $\road{k}$ that needs an additional agent carrying $\tk{j}$ (Line~\ref{ln:maoj_identify_next_road}). It moves $\ag{i}$ to the tail $\tail{}(\road{k})$ of $\road{k}$ (Line~\ref{ln:move_agent_to_exit_road}). It adds $\ag{i}$ to the set $\canChgTk{}$ (Line~\ref{ln:update_canChgTk}) since it is on a new road and updates $\inBM{}'$ (Line~\ref{ln:update_inB}) and $\arrivalT{}$ (Line~\ref{ln:update_arrivalT}).

\begin{algorithm}[t]
\small
\caption{$\moveAgentsOnRoad{}(\road{h}, t, \T{})$}
\label{alg:move_agent_on_road}
\begin{algorithmic}[1]
\For{$\cl{j} \in \reversePath{}(\road{h})$}
\label{ln:next_cell_on_road}
  \State $\ag{i} \gets \agentAt{}(\cl{j}, t)$
  \label{ln:maor_get_ai_cell}
  \If{$\ag{i} \neq \nullVal{}$}
    \If{$\exit{}(\cl{j}) = \head{}(\road{h}) \wedge \arrivalT{}(\ag{i}) = \T{}$}
    \label{ln:arrived_with_epoch_check}
        \State $\at{}(\cl{j}, t+1) \gets \ag{i}$
    \label{ln:agent_ai_waits_to_avoid_double_advance}
    \ElsIf{$\at{}(\exit{}(\cl{j}), t+1) = \nullVal{}$}
      \label{ln:is_exit_cell_unoccupied}
      \State $\at{}(\exit{}(\cl{j}), t+1)\gets \ag{i}$
      \label{ln:agent_ai_moves}
    \Else
      \State $\at{}(\cl{j}, t+1)\gets \ag{i}$
      \label{ln:agent_ai_waits}
    \EndIf
  \EndIf
\EndFor
\end{algorithmic}
\end{algorithm}

\para{\moveAgentsOnRoad{}.} $\TSTGEN{}$ then moves each agent $\ag{i}$ on a road $\road{h}$ with the function \moveAgentsOnRoad{} (Algorithm~\ref{alg:move_agent_on_road}). Let cell $\cl{j}$ be the cell that $\ag{i}$ is on (Line~\ref{ln:maor_get_ai_cell}). Let $\exit{}(\cl{j})$ be cell $\cl{j}$'s exit cell if cell $\cl{j}$ is a road cell. If agent $\ag{i}$ is at the head $\head{}(\road{h})$ of road $\road{h}$ and it entered $\road{h}$ on the current epoch, it waits on $\cl{j}$ for the next epoch to begin (Lines~\ref{ln:arrived_with_epoch_check}-\ref{ln:agent_ai_waits_to_avoid_double_advance}). Otherwise, if the cell $\exit{}(\cl{j})$ is currently unoccupied on timestep $t+1$, $\ag{i}$ moves onto $\exit{}(\cl{j})$ (Lines~\ref{ln:is_exit_cell_unoccupied}-\ref{ln:agent_ai_moves}). If it is, $\ag{i}$ waits on $\cl{j}$ (Line~\ref{ln:agent_ai_waits}).

Let the function $\reversePath{}(\road{h})$ return a list of the cells in road $\road{h}$ starting at its head and ending at its tail. We process the agents on a road $\road{h}$ in reverse order of their distance from its head (Line~\ref{ln:next_cell_on_road}). Consequently, whenever we process an agent $\ag{i}$ on a cell $\cl{j}$ on road $\road{h}$, we know if an agent in front of $\ag{i}$ on $\road{h}$ needs to wait at the cell $\exit{}(\cl{j})$.



\begin{algorithm}[t]
\small
\caption{\depositToken{}($\allM{}, t$)}
\label{alg:deposit_token}
\begin{algorithmic}[1]
\For{$\m{h} \in \allM{}$ s.t. $\neg \isSourceM{}(\m{h})$}
  \State $\ag{i} \gets \agentAt{}(\inputCl{}(\m{h}), t+1)$
  \If{$\ag{i} \neq \nullVal{} \wedge \agTk{}(\ag{i}, t) \neq \tk{0} \wedge \ag{i} \in \canChgTk{}$}
  \label{ln:check_if_agi_exists}
  \State $\tk{j} \gets \agTk{}(\ag{i}, t)$
\If{$\dpF{}'(\m{h}, \arrivalT{}(\ag{i}), \tk{j}) > 0$}
\label{ln:ai_has_token_for_mh}
  \State $\agTk{}(\ag{i}, t+1) \gets \tk{0}$
  \label{ln:deposit_token_instr}
  \State $\bIn{}(\m{h}, t+1) \gets \bIn{}(\m{h}, t+1) \cup \{\{\tk{j}\}\}$
  \label{ln:adjust_input buffer}
  \State $\canChgTk{} \gets \canChgTk{} \setminus \{\ag{i}\}$
  \label{ln:rm_from_canChgTk}
  \State $\dpF{}'(\m{h}, \arrivalT{}(\ag{i}), \tk{j}) {\gets} \dpF{}'(\m{h}, \arrivalT{}(\ag{i}), \tk{j}) {-} 1$
  \label{ln:adjust_deposit_counter}
\EndIf
\EndIf    
\EndFor
\end{algorithmic}
\end{algorithm}

\para{\depositToken{}.} $\TSTGEN{}$ then checks if each agent $\ag{i}$ on a machine $\m{h}$'s input cell $\inputCl{}(\m{h})$ carrying a non-null token $\tk{j}$ should deposit its token using the function \depositToken{} (Algorithm~\ref{alg:deposit_token}). If (a) agent $\ag{i}$ is allowed to deposit its token (Line~\ref{ln:check_if_agi_exists}) and (b) machine $\m{h}$ needs an additional agent which entered its current road on epoch $\arrivalT{}(\ag{i})$ to deposit a copy of token $\tk{j}$ (Line~\ref{ln:ai_has_token_for_mh}), agent $\ag{}{i}$'s token is placed in $\m{h}$'s input buffer (Lines~\ref{ln:deposit_token_instr}-\ref{ln:adjust_input buffer}). The variables $\canChgTk{}$ and $\dpM{}'$ are then updated (Lines~\ref{ln:rm_from_canChgTk}-\ref{ln:adjust_deposit_counter}).

\para{\pickupToken{}.} $\TSTGEN{}$ instructs agents to pick up tokens using the function \pickupToken{} (Algorithm~\ref{alg:TS_ACES}, Line~\ref{ln:pickup_token}). This function operates analogously to \depositToken{}.

\para{\keepToken{}.} If an agent neither picks up nor deposits a token, the token that it is holding does not change (Line~\ref{ln:keep_token}).

Finally, $\TSTGEN{}$ converts the occupancy vector $\atM{}(t+1)$ back into a position vector $\agClM{}(t+1)$ (Line~\ref{ln:genPositionVec}).

\para{\initializeSF{}.} \TSACES{} determines the smart factory's state on timestep $t=0$ using the function $\initializeSF{}$. Each agent is initialized in a queue at the head of a road. There are $\outB{}(\road{i}, 0, \tk{j})$ agents in the queue at the head of road $\road{i}$ carrying the token $\tk{j} \in \allTk{} \cup \{\tk{0}\}$. Agents do not pick up or deposit tokens until they have passed through an intersection for the first time. Each machine's input buffer is initialized with every token that it consumes in a single cycle of epochs. Since every one of these tokens will be replaced by the start of the next cycle, a machine will never run out of tokens. We initialize each machine's output buffer with every token that it emits in a single cycle for similar reasons.

\section{Analysis}

A solution to the SFE problem is complete if, given enough time, it can solve any SFE
instance.

\begin{theorem}
    \TSACES{} is complete.
\end{theorem}

\para{Proof.} By construction, any traffic system based embedding can be converted into a cell based embedding. \TSACES{} is therefore complete if it can construct a traffic system based embedding for any SFE instance. A smart factory's layout graph is strongly connected. There is thus a traffic system based embedding $\sol{}$ for any SFE instance whose transport plan uses a single agent. We will show that \TSACES{} will find $\sol{}$ if no better embedding exists. 

Let $\singleAgentPlan{}$ be the finite sequence of roads that $\sol{}$'s agent traverses every cycle. The \TSMILP{} is always run with a large enough epoch length to allow an agent on any of a junction's input roads to move to any of its output roads as long as there are not other agents on its input roads. Therefore, the \TSMILP{} can construct an embedding with a single agent that traverses the sequence of roads $\singleAgentPlan{}$ if it is run with $|\singleAgentPlan{}|$ epochs. If $\tsPlanner{}$ is given enough time, the \TSMILP{} will be run with $|\singleAgentPlan{}|$ epochs. $\square$

\begin{table*}[t]
\begin{center}
\begin{tabular}{llllllll}
\toprule
Scenario Name                                   & Machines &ACES Thr. &TS-ACES Thr. &ACES Runt.  &TS-MILP Runt. &$\TSTGEN{}$ Runt. &TS-ACES Agents Used\\\midrule
Contact Lens Small~\cite{contact_lens_production}  %
& 24   &2    &1.14  &60  &0.31  &0.004  &102\\
Contact Lens Large~\cite{contact_lens_production}  %
& 107  &2.14 &4.71  &60  &1.92  &0.01  &430\\
Drug Synthesis Small~\cite{API_generation}     & 18   &0.2 &0.095 &60  &11.6 &0.003 &44\\
Drug Synthesis Large~\cite{API_generation}     & 108   &N/A &0.5 &60  &60 &0.02 &222\\
Hard Candy Small~\cite{hard_candy_production}&8   &0.25 &0.14 &60 &0.16 &0.008 &28\\
Hard Candy Large~\cite{hard_candy_production}&104 &0.8 &1.86 &60 &60 &0.02 &368

\end{tabular}
\caption{Evaluation Results}
\vspace{-12mm}
\end{center}
\label{tbl:evaluation_results}
\end{table*}

\section{Evaluations}
Our evaluations investigate the following questions:
\begin{itemize}
    \item How do \TSACES{}'s solution quality and runtime compare to those of ACES~\cite{ACES}, a state-of-the-art SFE solver?
    \item Can the \TSMILP{} and $\TSTGEN{}$ scale to factories with more than a hundred machines  based on realistic scenarios?
\end{itemize}

\para{Implementation.} \TSACES{} is written in Python 3.11. The \TSMILP{} is expressed using the Gurobipy library and solved using the Gurobi MILP solver~\cite{gurobi}. 

\para{Experimental Hardware.} Our evaluations were performed on a 3.2 GHz, 8 Core AMD Ryzen 5800H CPU with 14 GB of RAM running Ubuntu 20.04.6 LTS.

\para{Methodology.} \TSACES{} is benchmarked on 6 scenarios taken from the food and pharmaceutical manufacturing industry. Companies in these industries frequently want to manufacture different versions of a product. A drug maker, for instances, frequently wants to manufacture different sizes of pill. As a result, these industries benefit from flexible manufacturing. Table II shows the number of machines in each scenario. Each of the contact lens, drug synthesis and hard candy scenarios have 6, 8 and 8 processes.  The solvers were allowed to use at most 1000 agents in their solutions. Roads contained at most one input or output cell. Each solver was given a maximum runtime of 60 seconds. TS-ACES was terminated if it failed to improve on its best solution after increasing $\numEpochs{}$ or $\epochLen{}$ $\gamma=2$ consecutive times. TS-ACES increased $\epochLen{}$ by $\delta=1$ each run.

\para{Results.} Table II shows ACES's and TS-ACES's throughput, ACES's, the \TSMILP{}'s and $\TSTGEN{}$'s runtime, and the number of agents that TS-ACES used on each of our 6 scenarios. Runtimes are measured in seconds. ACES outperforms \TSACES{} on each of its small scenarios, since its MILP does not have the traffic system based limitations that the \TSMILP{} does. ACES, however, generates less than half of \TSACES{}'s throughput on Contact Lens Large and Hard Candy Large and could not solve Drug Synthesis Large within 60 seconds. The \TSMILP{} can generate solutions for SFE instances with over 100 machines in 60 seconds. Despite the \TSMILP{} producing solutions that use 430, 222 and 368 agents, $\TSTGEN{}$ 0.01, 0.02 and 0.02 seconds to run (averaged over 100 runs). 
\TSACES{} is able to scale to very large SFE instances based on real industrial scenarios.

\section{Conclusion}

In this paper, we addressed the Smart Factory Embedding (SFE) problem with \TSACES{}, the Traffic System based Anytime Cyclic Embedding Solver. We analyze \TSACES{} and show that it is complete. In future work, we hope to scale \TSACES{} further. We also hope to find ways to automatically generate smart factory traffic system layouts.

\section*{Acknowledgement}

The research at the University of California, Irvine and University of
Southern California was supported by the National Science Foundation
(NSF) under grant numbers 2434916, 2346058, 2321786, 2121028, and
1935712 as well as gifts from Amazon Robotics. The views and conclusions
contained in this document are those of the authors and should not be
interpreted as representing the official policies, either expressed or
implied, of the sponsoring organizations, agencies, or the U.S.  government.

\bibliographystyle{IEEEtran}
\bibliography{IEEEabrv,references}

\end{document}